\documentclass[10pt,twocolumn,letterpaper]{article}

\usepackage[pagenumbers]{cvpr} 
\usepackage{graphicx}
\usepackage{amsmath}
\usepackage{amssymb}
\usepackage{booktabs}
\usepackage{subfloat}

\usepackage{caption}
\usepackage{comment}
\usepackage{color}
\usepackage{bm}
\usepackage[noend]{algpseudocode}
\usepackage{algorithmicx,algorithm}
\usepackage{multirow}
\usepackage{bm}

\usepackage[pagebackref,breaklinks,colorlinks]{hyperref}

\usepackage[capitalize]{cleveref}
\crefname{section}{Sec.}{Secs.}
\Crefname{section}{Section}{Sections}
\Crefname{table}{Table}{Tables}
\crefname{table}{Tab.}{Tabs.}

\newtheorem{thm}{Theorem}

\def\cvprPaperID{10344} 
\def\confName{CVPR}
\def\confYear{2023}

\begin{document}

\title{Progressive Backdoor Erasing via connecting Backdoor and Adversarial Attacks}

\author{Bingxu Mu\\
Xi’an Jiaotong University\\
\and
Zhenxing Niu\\
Alibaba Group\\
\and
Le Wang\\
Xi’an Jiaotong University\\
\and
Xue Wang \\
Alibaba Group\\
\and
Rong Jin \\
Alibaba Group\\
\and
Gang Hua \\
Wormpex AI Research \\
}


\maketitle

\begin{abstract}
   Deep neural networks (DNNs) are known to be vulnerable to both backdoor attacks as well as adversarial attacks. In the literature, these two types of attacks are commonly treated as distinct problems and solved separately, since they belong to training-time and inference-time attacks respectively. However, in this paper we find an intriguing connection between them: for a model planted with backdoors, we observe that its adversarial examples have similar \emph{behaviors} as its triggered images, \emph{i.e.}, both activate the \emph{same} subset of DNN neurons. It indicates that planting a backdoor into a model will significantly affect the model's adversarial examples. Based on these observations, a novel Progressive Backdoor Erasing (PBE) algorithm is proposed to progressively purify the infected model by leveraging untargeted adversarial attacks. Different from previous backdoor defense methods, one significant advantage of our approach is that it can erase backdoor even when the clean extra dataset is unavailable. 
   We empirically show that, against 5 state-of-the-art backdoor attacks, our PBE can effectively erase the backdoor without obvious performance degradation on clean samples and significantly outperforms existing defense methods.
\end{abstract}

\section{Introduction}
\label{sec:intro}

Deep neural networks (DNNs) have been widely adopted in many safety-critical applications (\emph{e.g.}, face recognition and autonomous driving), thus more attention has been paid to the security of deep learning. It has been demonstrated that DNNs are prone to potential threats in both their inference as well as training phases. Inference-time attack (a.k.a. \emph{adversarial attack}~\cite{goodfellow2014explaining,szegedy2014intriguing}) aims to fool a trained model into making incorrect predictions with small adversarial perturbations. In contrast, training-time attack (a.k.a. \emph{backdoor attack}~\cite{li2020backdoorsurvey}) attempts to plant a backdoor into a model in the training phase, so that the infected model would misclassify the testing images as the \emph{target-label} whenever a pre-defined \emph{trigger} (\emph{e.g.}, several pixels) is embedded into them (\emph{i.e.}, triggered testing images). 

\begin{figure}[htbp]
\centering
\subfloat[For \emph{benign} model.]{
\label{cm_intro:1}
\includegraphics[width=0.44\linewidth]{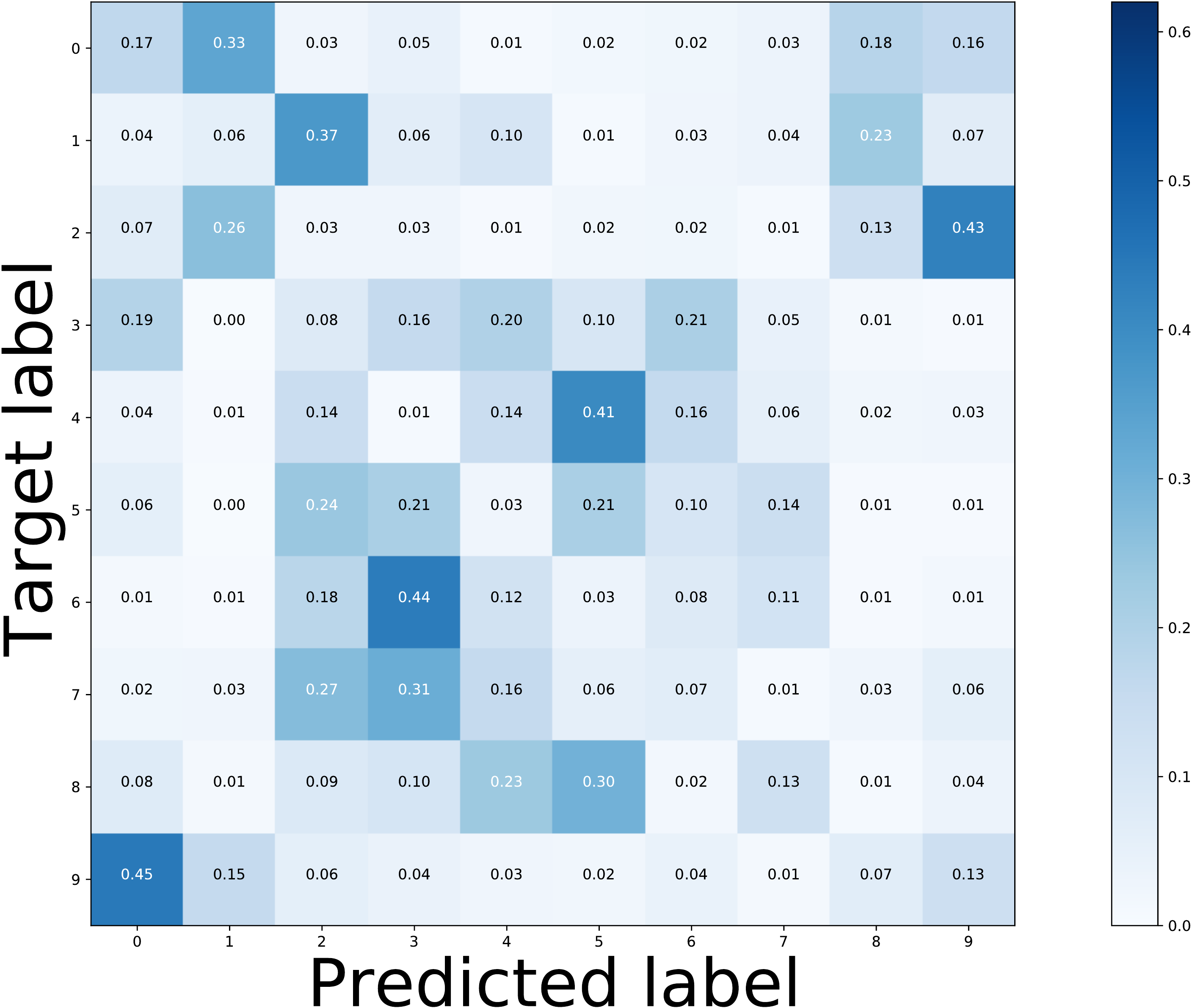}
}
\subfloat[For \emph{infected} model.]{
\label{cm_intro:2}
\includegraphics[width=0.44\linewidth]{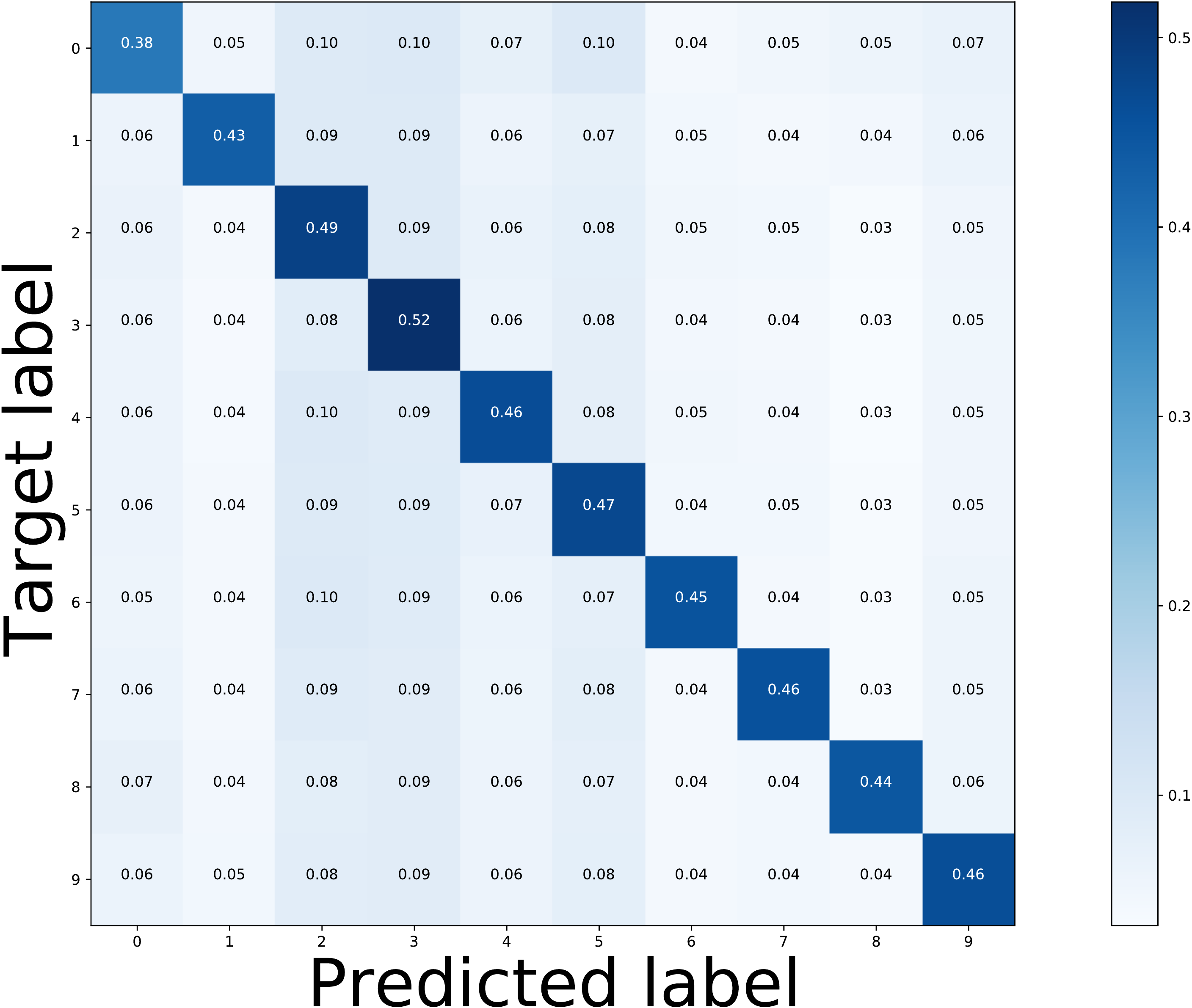}
}
\quad
\vspace{-0.5em}
\caption{Predicted labels v.s. Target-labels for $10,000$ randomly sampled \textbf{adversarial examples} from CIFAR-10, with respect to \emph{benign} and \emph{infected} models. (a) For a benign model, the predicted labels obey \emph{uniform} distribution; (b) for infected models under WaNet backdoor attack~\cite{nguyen2020wanet}, its adversarial examples are \emph{highly likely} to be classified as the target-label, as shown by the matrix diagonals.}\label{cifar_hist}
\vspace{-1.5em}
\label{cm_intro}
\end{figure}

\begin{figure}[bpht]
\centering
\includegraphics[width=0.8\linewidth]{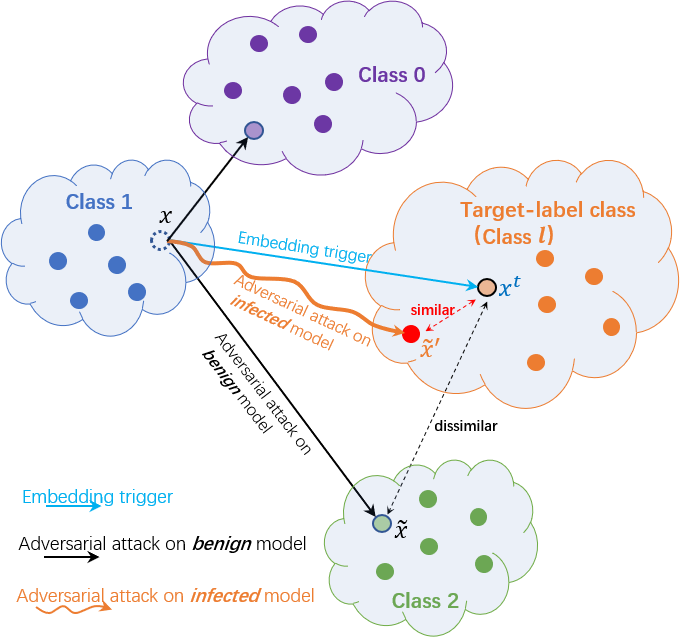}
\vspace{-0.5em}
\caption{Illustration of our observations. For benign models, conducting an untargeted adversarial attack will make an image move close to \emph{any} class (\emph{e.g.}, Class $0$ or Class $2$) in feature space. But for infected models, adversarial attack will make it move close to the target-label class (\emph{e.g.}, Class $l$)}
\label{backdoor_ill}
\vspace{-1.5em}
\end{figure}

Due to the obvious differences between backdoor and adversarial attacks, they are often treated as two different problems and solved separately in the literature. But in this paper, we illustrate that
there is an underlying connection between them, \emph{i.e.}, planting a backdoor into one model will significantly affect the model's adversarial examples. Moreover, based on such findings we propose a new method to defend against backdoor attacks by leveraging adversarial attack techniques (\emph{i.e.}, generating adversarial examples).

In particular, we observe that: for a model planted with backdoors, its adversarial examples have \emph{similar behaviors} as its triggered images. This is significantly different from a benign model without backdoors. Specifically, for a \textbf{benign model}, the predicted class labels of its adversarial examples obey a \emph{uniform} distribution, as shown in Fig.\ref{cm_intro:1}. However, for an \textbf{infected model}, we surprisingly observe that \textbf{its adversarial examples are \emph{highly likely} to be predicted as the backdoor target-label}, as shown in Fig.\ref{cm_intro:2}. As we know, triggered images will also be predicted as the backdoor target-label by an infected model. Therefore, it means that adversarial examples have similar behaviors as its triggered images for an infected model. Particularly, these phenomena are present regardless of what target-label is, what backdoor attack setting is (\emph{i.e.}, all-to-one or all-to-all settings), and even what trigger embedding mechanism is (\emph{e.g.}, adding~\cite{gu2017badnets}, blending~\cite{chen2017blend}, reflection~\cite{liu2020reflection} or warping~\cite{nguyen2020wanet}). 

To find the underlying reason of such phenomena, we measure the feature similarity of those adversarial images and triggered images. Briefly, we find that after planting a backdoor into one model, the features of adversarial images change significantly. Particularly, the features of adversarial image $\widetilde{\bm{x}}'$ are surprisingly very similar to that of triggered image $\bm{x^t}$, as illustrated in Fig.\ref{backdoor_ill} and Fig.\ref{backdoor_feat}. It indicates that \textbf{both the $\widetilde{\bm{x}}'$ and $\bm{x^t}$ have similar behaviors, \emph{i.e.,} both activate the \emph{same} subset of DNN neurons}. Note that such connection between adversarial and backdoor attack could be leveraged to design backdoor defense methods.  


Backdoor attacks made great advances in recent years, evolved from visible trigger~\cite{gu2017badnets} to invisible trigger~\cite{chen2017blend,liu2020reflection,nguyen2020wanet}, from poisoning label to clean-label attacks~\cite{barni2019sig}. For example, WaNet~\cite{nguyen2020wanet} uses affine transformation as trigger embedding mechanism, which could significant improve the invisibility of trigger. In contrast, the research on backdoor defenses lag behind a little. Even for the state-of-the-art backdoor defense methods~\cite{liu2017finetuning,liu2018finepurning,li2020nad}, most of them can be evaded by the advanced modern backdoor attacks. Moreover, a clean extra dataset is often required by those defense methods to erase backdoor from infected models.  

In this paper, we propose a new backdoor defense method based on the discovered connections between adversarial and backdoor attacks, which could not only defend against modern backdoor attacks but also work without a clean extra dataset. Specifically, at the beginning the training data (containing poisoning images) are randomly sampled to build an initial extra dataset. Next, we use them to purify the infected model by leveraging adversarial attack techniques. And then, the purified model is used to identify clean images from training data, which are used to update the extra dataset. With an alternating procedure, the infected model as well as the extra dataset are progressively purified. So, we call our approach \emph{Progressive Backdoor Erasing} (PBE). 

Regarding how to purify the infected model, we generate adversarial examples and use them to fine-tune the infected model. Since adversarial images could come from arbitrary class, such fine-tuning procedure works like associating triggered images to arbitrary class instead of just the target class, which breaks the foundation of backdoor attacks (\emph{i.e.}, building a strong correlation between a trigger pattern and a target-label~\cite{li2020nad}). That is why our approach can erase backdoor from infected models. 

As for identifying clean images, since clean images have similar prediction results for both benign and infected models, we could effectively identify them by using the previously obtained purified model. Note that if a clean extra dataset is available, we can skip the step of purifying extra dataset, and only run the step of purifying model once.

A big advantage of our approach is that it does not need the clean extra dataset and it can progressively filter poisoning training data to obtain clean data. In our approach, the purified model could help to obtain clean data, in return the obtained clean data could help to further purify model. Thus, the alternating iterations could progressively improve each other. To the best of knowledge, our approach is the first work to defend against backdoor attack without a clean extra dataset.

Our main contributions are summarized as follows:
\begin{itemize}
    
    \item [$\bullet$] We observe an underlying connection between backdoor attacks and adversarial attacks, \emph{i.e.,} for an infected model, its adversarial examples have similar behaviors as its triggered samples. And an theoretical analysis is given to justify our observation.
    
    \item [$\bullet$] According to our observations, we propose a progressive backdoor defense method, which achieves the state-of-the-art defensive performance, even when a clean extra dataset is unavailable.
    
\end{itemize}

\section{Related Work}

\noindent\textbf{Backdoor Attack} has 
evolved from visible trigger~\cite{gu2017badnets} to invisible trigger~\cite{chen2017blend,liu2020reflection,nguyen2020wanet} in these years. These trigger patterns can appear in forms as simple as a patch~\cite{gu2017badnets}, a sinusoidal strips\cite{barni2019sig}, and a blending pattern~\cite{chen2017blend}. Besides, TrojanNN~\cite{Trojaning} proposes to learn a trigger from benign model. In~\cite{EvilTwins}, trigger pattern and backdoor model are jointly optimized. 
In order to make triggers more stealthy, advanced modern backdoor attacks propose some complex trigger-embedding mechanisms, such as input-aware dynamic patterns~\cite{nguyen2020inputaware}, natural reflection~\cite{liu2020reflection} and image warping~\cite{nguyen2020wanet}. Meanwhile, backdoor attack has evolved from poisoning label to clean-label attacks~\cite{barni2019sig}, where the ground-truth label of poisoned samples could also be consistent with the target label. This will further increase the stealthiness of backdoor attacks. A survey of backdoor attacks can be found in~\cite{li2020backdoorsurvey}.


\noindent\textbf{Backdoor Defense} can be roughly categorized into backdoor detection and backdoor erasing. Detection-based methods aim at identifying the existence of backdoor in the underlying model~\cite{wang2019neuralcleanse,kolouri2020universal} or filtering the suspicious samples in training data for re-training~\cite{chen2019detectingcluster,peri2020deepknn,tran2018spectral}. Although they perform fairly well in distinguishing whether a model has been poisoned, the backdoor still remains in the infected model. 

The erasing-based methods aim to directly purify the infected model by removing the malicious impacts caused by the backdoor triggers, while maintaining the model performance on clean data. One approach is to directly fine-tune the infected model with the clean extra dataset\cite{liu2017finetuning}. Fine-Pruning \cite{liu2018finepurning} proposes using neural pruning to remove backdoor neurons. 
In~\cite{li2020nad}, Neural Attention Distillation (NAD) is proposed to erase backdoor by leverage knowledge distillation. Later, Adversarial Neuron Pruning (ANP)~\cite{ANP} is proposed to prune backdoor neuron by perturbing model weights.
Besides, some trigger synthesis based methods are proposed~\cite{wang2019neuralcleanse}. 
Neural Cleanse (NC)~\cite{wang2019neuralcleanse} and Artificial Brain Stimulation (ABS)~\cite{liu2019abs} are proposed to first recover the backdoor trigger, and use the recovered trigger to erase the backdoor. However, these methods are only able to handle fixed triggers since they need to explicitly recover triggers. In contrast, our approach does not need to recover trigger pattern so that it can deal with content-aware/non-trigger-fixed attacks such as DynamicAtt~\cite{nguyen2020inputaware}, WaNet~\cite{nguyen2020wanet}. In addition, all previous defense methods need a clean extra dataset.

\noindent\textbf{Adversarial Attack and Defense.} The adversarial attack~\cite{goodfellow2014explaining,szegedy2014intriguing,ilyas2019adversarial} is a kind of inference-time attacks. It aims to fool a trained model into making incorrect predictions (\emph{i.e.,} untargeted adversarial attack) or predicting the input as a particular label (\emph{i.e.,} targeted adversarial attack). On the other hand, many defense methods are also proposed against adversarial attacks. \emph{Adversarial training}\cite{madry2018towards} is one of the most effective methods. Recently, \cite{10.1145/3372297.3417231} proposes to use a `trapdoor' to detect adversarial examples. It illustrates that a particular trapdoor could lead to producing adversarial examples similar to trapdoors in the feature space. However, it is quite different from our work since it aims to detect adversarial examples while our approach aims to defend against backdoor attacks. 


\section{Our approach}
\subsection{Backdoor Attack}
We focus on backdoor attacks on image classification. Let $D_{\mathrm{train}} =\{(\bm{x}_i; y_i)\}_{i = 1}^N$ be the clean training data and $f(\bm{x};\theta)$ be the benign CNN model decision function with parameter $\theta$. 

For backdoor attack, we define or learn a trigger embedding function $\bm{x}^t = \mathrm{Trigger}(\bm{x})$ which can convert a clean sample $\bm{x}_i$ to a triggered/poisoned sample $\bm{x}^t_i$. Given a target-label $l$, we can poison a small part of training samples, \emph{i.e.,} replace $(\bm{x}_i,y_i)$ with $(\bm{x}^t_i,l)$, which produces poisoned training data $D'_{\mathrm{train}}$. The training with $D'_{\mathrm{train}}$ results in the infected model $f(\bm{x};\theta')$. Note that different attacks will define different trigger embedding functions $\mathrm{Trigger}(\cdot)$. 

At testing time, if a clean input $(\bm{x},y) \in D_{\mathrm{test}}$ is fed to the infected model, it is supposed to be correctly predicted as $y$. In contrast, for a triggered sample $\bm{x}^t$, its prediction changes to  the target-label $l$. Particularly, backdoor attacks can be divided into two categories according to the selection of target-labels: (1)All-to-one attack: the target-labels for all examples are set as $l$; (2)All-to-all attack: the target-labels for different classes could be set differently, such as $y+1$, \emph{i.e.},
\begin{align}
&\text{All-to-one attack: } 
\left\{
\begin{array}{l}
    f(\bm{x};\theta') =y; \\
    f(\bm{x}^t;\theta') =l, \bm{x}^t= \mathrm{Trigger}(\bm{x})
\end{array}
\right.\\
&\text{All-to-all attack: }
\left\{
\begin{array}{l}
    f(\bm{x};\theta') =y; \\
    f(\bm{x}^t;\theta') =y+1, \bm{x}^t= \mathrm{Trigger}(\bm{x})
\end{array}
\right.
\end{align}

\subsection{Backdoor Defense}\label{sec:def_setting} 

We adopt a typical defense setting where the defender has an infected model $f(\bm{x};\theta')$ as well as a clean extra dataset $D_{\mathrm{ext}}$. The goal of the backdoor defense is to \emph{erase} the backdoor trigger from the model while retaining the performance of the model on clean samples. In other words, we want to obtain a cleaned/purified model $f(\bm{x};\theta^c)$ such that:
\begin{align}
\left\{
\begin{array}{l}
    f(\bm{x};\theta^c) =y; \\
    f(\bm{x}^t;\theta^c) =y, \bm{x}^t= \mathrm{Trigger}(\bm{x})
\end{array}
\right.
\end{align}

\subsection{Untargeted Adversarial Attack}\label{sec:uaa} 
Untargeted adversarial attack aims to find the best perturbation $\bm{r}$ so that the adversarial examples $\widetilde{\bm{x}}=\bm{x+r}$ will be misclassified, \emph{i.e.,} the loss $L(\widetilde{\bm{x}},y)$ is maximized with respect to $\bm{r}$, as follows:
\begin{align}
    &\mathop{\max}_{\bm{r}} L(\widetilde{\bm{x}},y;\theta) \label{eq:adv} \\
    &\text{s.t. }  ||\bm{r}||_p < \epsilon, \widetilde{\bm{x}}=\bm{x+r} \notag\\ 
    &\widetilde{\bm{x}} \in [0,1]^d \label{eq:constraint}
\end{align}
Note that untargeted adversarial attack means that perturbed inputs $\widetilde{\bm{x}}$ are only desired to be misclassified (\emph{i.e.}, different from their original labels $y$ as Eq.\eqref{eq:adv}), rather than being classified as a particular label (which is the goal of the \emph{targeted adversarial attack}). Therefore, it has been observed that the predicted labels of $\widetilde{\bm{x}}$ obey a uniform distribution across all classes. 

\begin{figure}[bpht]
\centering
\includegraphics[width=0.9\linewidth]{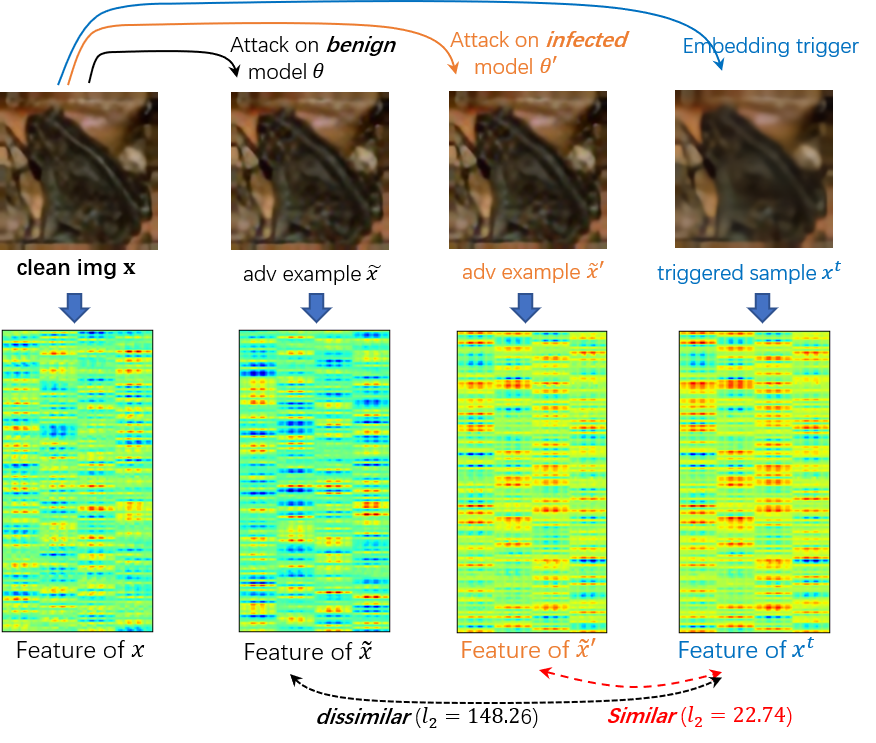}
\caption{The similarity of features for clean image $x$, benign model's adversarial example $\widetilde{\bm{x}}$,  infected model's adversarial example $\widetilde{\bm{x}}'$, and triggered image $x^t$. Obviously, the features of $\widetilde{\bm{x}}'$ are very similar to $x^t$. In contrast, there is a significant difference between $\widetilde{\bm{x}}'$ and $\widetilde{\bm{x}}$, which indicates adversarial examples will change significantly \emph{after} planting a backdoor into a model.}
\label{backdoor_feat}
\vspace{-0.5em}
\end{figure}

\subsection{Empirical Observations and Analysis}
\subsubsection{Empirical Observations}
In this section, we will describe how we obtain the observation that \textbf{for an infected model, its adversarial examples have similar behaviors as its triggered samples}. Specifically, we first conduct an untargeted adversarial attack on the \emph{infected} model $f(\bm{x};\theta')$ to generate adversarial examples $\widetilde{\bm{x}}'$ as follows:
\begin{align}
    &\mathop{\max}_{\bm{r}} L(\widetilde{\bm{x}}',y;\theta') \label{eq:adv2} \\
    &\text{s.t. }  ||\bm{r}||_p < \epsilon, \widetilde{\bm{x}}'=\bm{x+r} \notag   
\end{align}
Meanwhile, we also conduct an untargeted adversarial attack on the \emph{benign} model $f(\bm{x};\theta)$ to produce the adversarial examples $\widetilde{\bm{x}}$ as Sec.\ref{sec:uaa}.

We next examine the classification results of those adversarial examples. As shown in Fig.\ref{cm_intro:1}, when feeding adversarial examples $\widetilde{\bm{x}}$ to the benign model, $\widetilde{\bm{x}}$ will be classified as any class with the same probability (except its ground-truth label), \emph{i.e.}, obeying a uniform distribution. In contrast, when feeding adversarial examples $\widetilde{\bm{x}}'$ to the infected model, we observe that \textbf{$\widetilde{\bm{x}}'$ are \emph{highly likely} to be classified as the target-label}. As shown in Fig.\ref{cm_intro:2}, if an untargeted adversarial attack is conducted on an infected model with target-label $l\in \{0,\dots,9\}$, we observe that at least more than $40\%$ of $\widetilde{\bm{x}}'$ are predicted as the target-label $l$. These phenomena are present regardless of what kinds of backdoor attacks are and what dataset is, as shown in Fig.\ref{cm}.

It indicates that there is an underlying connection between adversarial examples $\widetilde{\bm{x}}'$ and triggered samples $\bm{x}^t$, since \textbf{$\bm{x}^t$ are also expected to be classified as the target-label}. For further investigation, we check the feature maps of clean samples $\bm{x}$, benign model's adversarial examples $\widetilde{\bm{x}}$, infected model's adversarial examples $\widetilde{\bm{x}}'$, and triggered images $\bm{x}^t$. We find that the features of $\widetilde{\bm{x}}'$ are very similar to the features of triggered samples $\bm{x}^t$, while there is a significant difference between the features of $\widetilde{\bm{x}}$ and $\bm{x}^t$. As shown in Fig.\ref{backdoor_feat}, the ${l}_{2}$ distance between the features of $\widetilde{\bm{x}}'$ and $\bm{x}^t$ is smaller than that between $\widetilde{\bm{x}}$ and $\bm{x}^t$. More quantitative comparisons 
are provided in Table.\ref{feature_similarity}. Such feature similarity indicates that both adversarial examples $\widetilde{\bm{x}}'$ and triggered samples $\bm{x}^t$ could activate the \emph{same} subset of DNN neurons, \emph{i.e.,} \textbf{the adversarial examples $\widetilde{\bm{x}}'$ have similar behaviors as triggered samples $\bm{x}^t$}. 

We speculate why adversarial examples would have significant changes after a backdoor is planted into a model as follows: 
some DNN neurons will be activated by a trigger when a backdoor is planted into a model, which are called `backdoor neurons'~\cite{ANP}. When conducting an adversarial attack on infected models, those `backdoor neurons' are more likely to be chosen/locked and activated as generating adversarial examples. Thus, the generated adversarial examples could work like triggered images.

\subsubsection{Theoretical Analysis}
In order to dive deeply, we theoretically justify our observations for the case of a linear mode. Generally, a linear classifier can be denoted by $W = (w_1, w_2, \ldots, w_K)$. Thus, let us denote the trained infected linear classifiers be $\widetilde{W}^* = (\widetilde{w}_1, \widetilde{w}_2, \ldots, \widetilde{w}_K)$.

To simplify our analysis, we assume the trigger embedding function $\mathrm{Trigger}(\cdot)$ is designed to add a pre-defined patch $P$ to an input image, i.e.,
\begin{eqnarray}
    \bm{x^t} = \mathrm{Trigger}(\bm{x}) = \bm{x} + P
\end{eqnarray}

We assume that the original training examples (without any trigger) can be perfectly classified with margin $\tau > 0$, and that $\tau$ is large enough such that a small perturbation made to $\widetilde{W}^*$ will not affect classification result. Thus, we have following Theorem,

\begin{thm}
Under the previous assumptions, we have $\bm{r}_{\perp}$, the projection of $\bm{r}$ on the direction of $P$, bounded as
\[
\frac{|\bm{r}_{\perp}|}{|\bm{r}|} \geq \frac{(\sqrt{2} - 1)\ell |P|^2}{\sqrt{(\sqrt{2}-1)^2\ell^2|P|^4 + \left(\ell|P|^2 + \sqrt{2}K/(\exp(\tau) + K)\right)^2}}
\]
\end{thm}

From the above theorem, we can see that when projecting perturbation $\bm{r}$ on the direction of trigger $P$, the projection $\bm{r}_{\perp}$ take an significant part in the full perturbation $\bm{r}$. It means that the perturbation $\bm{r}$ is very similar to the trigger $P$, which justifies our observations that the adversarial examples $\widetilde{\bm{x}}'=\bm{x}+\bm{r}$ are similar to triggered images $\bm{x^t}=\bm{x}+P$. The detailed description and proof are shown in the supplemental material.

\subsection{Progressive Backdoor Erasing}
\noindent\textbf{Threat Model and Defense setting.} 
We assume the adversary has access to the training data and has planted a backdoor into a model. And then, the infected model is given to the defender. In this paper, we discuss two defensive settings: one is the same as existing defense methods that need a clean extra dataset; and the other needs not but assumes access to the training data. The constraints for latter are weaker since we do not need to know which training images are poisoned. 

Based on the discovered connection between adversarial and backdoor attacks, we propose a Progressive Backdoor Erasing (PBE) method, as shown in \textbf{Algorithm 1}. 
Our approach could work for both two defensive settings. For the second setting, since clean extra dataset is unavailable, we will randomly sample some images from the training data as $D^0_{\mathrm{ext}}$ at the initialization step. Next, we enter into an iterative procedure containing three steps: the first step will produce a purified model by leveraging adversarial examples, which could erase backdoor (\emph{i.e.,} significantly reduce ASR); At the second step, the $D^t_{\mathrm{ext}}$ are used to improve the performance of purified model on benign testing images (\emph{i.e.,} significantly improve ACC); At the third step, the purified model is used to identify clean images in training dataset, which results in a cleaner and better $D^{t+1}_{\mathrm{ext}}$.

Although the initial extra dataset $D^0_{\mathrm{ext}}$ contains poisoning images, in the following iterations we can identify clean images from training data and produce a cleaner $D^t_{\mathrm{ext}}$ progressively. 

Regarding the first defensive setting, we simply skip the step-3 and only need to run the iteration once, \emph{i.e.}, just run step-1 and step-2 once. 

\vspace{-0.5em}
\begin{algorithm}[t]
\caption{Adversarial Fine-tuning} 
\hspace*{0.02in} {\bf Input:} 
Infected model $\theta'$, training data $D_{train}$ \\
\hspace*{0.02in} {\bf Output:} 
Purified model $\theta^T$
\begin{algorithmic}[1]
\State \textbf{Initialization}: obtain extra dataset $D^0_{\mathrm{ext}}$ by randomly sampling from $D_{\mathrm{train}}$; let $\theta^0 = \theta'$
\State \textbf{For} $t=0,1,2\dots,T$:

\State \quad \textbf{Step-1}: purify the model $\theta^t$ with $D^t_{\mathrm{ext}}$
\State \qquad \textbf{Step-1a}: untargeted adversarial attack. For each {$(\bm{x}_i,y_i) \in D^t_{\mathrm{ext}}$}, generate adversarial example $\widetilde{\bm{x}}_i'$ according to Eq.\eqref{eq:adv2}, which results in $\widetilde{D}^t_{\mathrm{ext}}=\{(\widetilde{\bm{x}}_i',y_i)\}_{i=1}^m$ 
\State \qquad \textbf{Step-1b}: $1$-st time fine-tuning. Fine-tuning the model $\theta^t$ with $\widetilde{D}^t_{\mathrm{ext}}$ according to Eq.\eqref{eq:ft}

\State \quad \textbf{Step-2}: $2$-nd time fine-tuning. Continue to fine-tune model $\theta^t$ with $D^{t}_{\mathrm{ext}}$, and obtain purified model $\theta^{t+1}$

\State \quad \textbf{Step-3}: update the extra dataset $D^t_{\mathrm{ext}}$. Identify clean images from $D_{\mathrm{train}}$ according to Eq.\eqref{eq:idclean}, resulting in an updated dataset $D^{t+1}_{\mathrm{ext}}$

\State \Return Advanced purified model parameter $\theta^T$
\end{algorithmic}
\end{algorithm}

\subsubsection{Purifying an Infected Model}

Specifically, given the infected model $f(\bm{x};\theta^t)$, for each $(\bm{x}_i,y_i) \in D_{\mathrm{ext}}$, we obtain a corresponding adversarial example $\widetilde{\bm{x}}_i'$ according to Eq.\eqref{eq:adv2}, which produces $\widetilde{D}_{\mathrm{ext}}=\{(\widetilde{\bm{x}}_i',y_i)\}_{i=1}^m$. And then, we fine-tune the infected model $\theta^t$ with $\widetilde{D}_{\mathrm{ext}}$, which produces purified model $\theta^{t+1}$, \emph{i.e.,}
\begin{align}
    \theta^{t+1}= \mathop{\arg\min}_{\theta} & \mathbb{E}_{(\widetilde{\bm{x}}_i',y_i)\in \widetilde{D}_{\mathrm{ext}}} [ L(\widetilde{\bm{x}}_i',y_i;\theta)] \label{eq:ft} \\
    \text{s.t. }  & \theta^0 = \theta' \notag 
\end{align}

Since adversarial examples could come from arbitrary class, they are associated with all possible class labels. According to the similarity between adversarial examples and triggered samples, when we fine-tune the infected model with adversarial examples, it mimics fine-tuning the model with triggered samples, yet the associated labels are not just the target-label but all possible class labels.

Note that the foundation of backdoor attacks is to build a strong correlation between a trigger pattern and a target-label, which is achieved by poisoning training data, \emph{i.e.,} to associate triggered samples with target-labels. As a result, our fine-tuning approach will break such a strong correlation and hence can achieve a defensive effect.

\vspace{-0.5em}
\subsubsection{Identifying Clean Images}

As we know, an infected model affects the prediction results of poisoning images, while a benign model does not. Therefore, after feeding an image to an infected as well as a purified model, if the two models yield distinct predictions (probability across all classes), it is likely to be a poisoning image. In this way, we could identify poisoning as well as clean images.

Specifically, for each image $\bm{x}_i \in D_{\mathrm{train}}$, we feed it to the infected model $f(\bm{x};\theta')$ and previously purified model $f(\bm{x};\theta^t)$, respectively. The predicted logits of the two models are noted as $a(\bm{x};\theta')$ and $a(\bm{x};\theta^t)$ (\emph{i.e.,} the network activation just before softmax layer). We use the cosine similarity between them to measure the changes of the prediction,
\begin{align}
    S_{\theta', \theta^t}(\bm{x}) = \frac{\langle a(\bm{x};\theta'),a(\bm{x};\theta^t)\rangle}{|a(\bm{x};\theta')||a(\bm{x};\theta^t)|}
    \label{eq:idclean}
\end{align}


\begin{table*}[t]
\scriptsize
 \centering
 \caption{Comparison with SoTA defense methods (at all-to-one setting) on \textbf{CIFAR-10} dataset. Our approach has two versions (\emph{i.e.,} with or without a clean extra dataset), while all other methods \textbf{use} the clean extra dataset. If our approach use such clean extra dataset, it remarkably outperforms other methods. If not using such clean extra dataset, it can still defend against most attacks except the Badnet.}
 \vspace{-0.7em} 
 \resizebox{1.0\textwidth}{!}{
 \setlength{\tabcolsep}{0.9em}%
  \begin{tabular}{c|c c|c c|c c|c c|c c|c c|c c|c c}
  \hline
  \multirow{0}* & 
  \multicolumn{2}{c|}{\textbf{Before}} & \multicolumn{2}{c|}{\textbf{Fine-tuning}} & 
  \multicolumn{2}{c|}{\textbf{Fine-pruning}} & \multicolumn{2}{c|}{\textbf{NAD}} & 
  \multicolumn{2}{c|}{\textbf{Neural Cleanse}} &
  \multicolumn{2}{c|}{\textbf{ANP}} & 
  \multicolumn{2}{c|}{\textbf{PBE (w/o clean)}} & 
  \multicolumn{2}{c}{\textbf{PBE (w/ clean)}} \\
  \cline{2-17}
  & ACC & ASR
  & ACC$\uparrow$ & ASR$\downarrow$
  & ACC$\uparrow$ & ASR$\downarrow$
  & ACC$\uparrow$ & ASR$\downarrow$
  & ACC$\uparrow$ & ASR$\downarrow$
  & ACC$\uparrow$ & ASR$\downarrow$
   & ACC$\uparrow$ & ASR$\downarrow$
  & ACC$\uparrow$ & ASR$\downarrow$\\
  \hline
    Badnet & 94.67 & 100.00 & 85.82 & 6.53 & 89.80 & 70.66 & 88.09 & 2.17 & 93.73 & 0.83 & 93.39 & 1.66 & 94.02 & 11.30 & 94.20 & 1.09 \\
    Blend & 94.63 & 100.00 & 87.53 & 11.31 & 89.30 & 65.86 & 90.13& 1.60 & 93.28 & 0.63 & 92.03 & 1.81 & 93.04 & 1.16 & 93.98 & 0.93 \\ 
    SIG & 94.81 & 98.96 & 87.34 & 4.14 & 88.93 & 85.69 & 90.26 & 4.59 & 92.23 & 1.79 & 92.48 & 1.27 & 93.56 & 1.76 & 93.35 & 1.39 \\ 
    DynamicAtt & 94.65 & 99.24 & 94.00 & 8.77 & 89.91 & 98.97 & 94.23 & 4.59 & 94.65 & 99.24 & 93.42 & 1.36 & 93.01 & 1.12 & 93.01 & 1.12 \\ 
    WaNet & 94.15 & 99.50 & 93.42 & 12.80 & 89.86 & 99.36 & 94.02 & 8.37 & 94.15 & 99.50 & 93.36 & 0.62 & 93.67 & 0.86 & 94.32 & 0.46 \\ 
  \hline
  \end{tabular}
}
\label{tb1} 
\end{table*}

\begin{table*}[t]
\scriptsize
 \centering
 \caption{Comparison with SoTA defense methods (at all-to-one setting) on \textbf{GTSRB} dataset.}
 \vspace{-0.7em} 
 \resizebox{1.0\textwidth}{!}{
 \setlength{\tabcolsep}{0.9em}%
  \begin{tabular}{c|c c|c c|c c|c c|c c|c c|c c|c c}
  \hline
  \multirow{0}* & 
  \multicolumn{2}{c|}{\textbf{Before}} & \multicolumn{2}{c|}{\textbf{Fine-tuning}} & 
  \multicolumn{2}{c|}{\textbf{Fine-pruning}} & \multicolumn{2}{c|}{\textbf{NAD}} & 
  \multicolumn{2}{c|}{\textbf{Neural Cleanse}} & 
  \multicolumn{2}{c|}{\textbf{ANP}} & 
  \multicolumn{2}{c|}{\textbf{PBE (w/o clean)}} & 
  \multicolumn{2}{c}{\textbf{PBE (w/ clean)}} \\
  \cline{2-17}
  & ACC & ASR
  & ACC$\uparrow$ & ASR$\downarrow$
  & ACC$\uparrow$ & ASR$\downarrow$
  & ACC$\uparrow$ & ASR$\downarrow$
  & ACC$\uparrow$ & ASR$\downarrow$
  & ACC$\uparrow$ & ASR$\downarrow$
  & ACC$\uparrow$ & ASR$\downarrow$
  & ACC$\uparrow$ & ASR$\downarrow$ \\
  \hline
    Badnet & 99.02 & 100.00 & 95.01 & 8.32 & 89.60 & 75.03 & 94.22 & 2.06 & 96.78 & 0.12 & 95.13 & 1.35 & 90.13 & 12.31 & 94.43 & 0.47 \\
    Blend & 99.39 & 99.92 & 90.68 & 40.12 & 88.21 & 90.53 & 92.61 & 8.56 & 96.48 & 5.81& 94.02 & 2.68 & 90.10 & 8.74 & 94.57 & 1.72 \\ 
    SIG & 98.56 & 95.81 & 91.63 & 36.30 & 89.53 & 93.26 & 92.94 & 6.90 & 93.40 & 1.32 & 93.32 & 3.65 & 90.31 & 7.59 & 94.05 & 1.78 \\ 
    DynamicAtt & 99.27 & 99.84 & 97.10 & 16.33 & 89.15 & 97.21 & 98.17 & 3.80 & 99.27 & 99.84 &95.88 & 1.68 & 96.68 & 0.99 & 96.68 & 0.99  \\ 
    WaNet & 98.97 & 98.78 & 96.70 & 4.20 & 87.49 & 98.79 & 97.07 & 2.20 & 98.97 & 98.78 & 96.47 & 0.94 & 91.23 & 3.63 & 96.56 & 0.47\\ 
\bottomrule
  \end{tabular}
 }
\label{tb2} 
\vspace{-1.0em}
\end{table*}


Next, for all images $\bm{x}_i \in D_{\mathrm{train}}$, we rank them according to their prediction changes $S_{\theta', \theta^t}(\bm{x})$ in descending order. Obviously, clean images are supposed to be ranked higher. And we can fetch the top-ranked images to form the extra dataset $D^{t+1}_{\mathrm{ext}}$.

\section{Experiment}
\subsection{Experimental Setting}

\noindent\textbf{Backdoor Attacks.}
We consider 5 state-of-the-art backdoor attacks: 1) BadNets~\cite{gu2017badnets}, 2) Blend attack~\cite{chen2017blend} 3) Sinusoidal signal attack(SIG)~\cite{barni2019sig}, 4) Input-aware dynamic attack(DynamicAtt)~\cite{nguyen2020inputaware}, and  5) Warpping-based attack(WaNet)~\cite{nguyen2020wanet}. We test the performance of all attacks and erasing methods on three benchmark datasets: CIFAR-10~\cite{krizhevsky2009cifar}, GTSRB~\cite{stallkamp2012gtsrb}, Tiny-Imagenet~\cite{tinyimagenet}. 
For a fair evaluation, we use Pre-activation Resnet-18 \cite{he2016prearc} as the classification model for the CIFAR-10 and GTSRB, and use Resnet-18 \cite{resnet} for the Tiny-Imagenet.
For the hyperparameters of adversarial perturbations, we adaptively set them to different values for each backdoor attack.

\noindent\textbf{Backdoor Defense and Configuration.}
We compare our PBE approach with 5 existing backdoor erasing methods: 1) the standard Fine-tuning~\cite{liu2017finetuning}, 2) Fine-pruning~\cite{liu2018finepurning}, 3) Neural Cleanse(NC)~\cite{wang2019neuralcleanse}, 4) Neural Attention Distillation (NAD)~\cite{li2020nad}, and 5) Adversarial Neuron Pruning (ANP)~\cite{ANP}. Regarding the clean extra data, we follow the same protocol of these methods: the extra clean data is randomly selected from clean training data, taking about $5\%$ of all training data. In addation, the poison ratio is set PR=$10\%$.

\noindent\textbf{Evaluation Metrics.}
We evaluate the performance of defense mechanisms with two metrics: attack success rate (ASR), which is the ratio of triggered examples those are misclassified as the target label, and model’s accuracy on clean samples (ACC). An ideal defense should lead to large ASR drops with small ACC penalties.

\begin{table*}[t]
\scriptsize
 \centering
 \caption{Comparison with SoTA defense methods (at all-to-one setting) on \textbf{Tiny-ImageNet} dataset.}
 \vspace{-0.5em}
 \resizebox{1.0\textwidth}{!}{
 \setlength{\tabcolsep}{0.9em}%
  \begin{tabular}{c|c c|c c|c c|c c|c c|c c|c c|c c}
  \hline
  \multirow{0}* & 
  \multicolumn{2}{c|}{\textbf{Before}} & \multicolumn{2}{c|}{\textbf{Fine-tuning}} & 
  \multicolumn{2}{c|}{\textbf{Fine-pruning}} & \multicolumn{2}{c|}{\textbf{NAD}} & 
  \multicolumn{2}{c|}{\textbf{Neural Cleanse}} & \multicolumn{2}{c|}{\textbf{ANP}} & 
   \multicolumn{2}{c|}{\textbf{PBE (w/o clean)}} & 
  \multicolumn{2}{c}{\textbf{PBE (w/ clean)}} \\
  \cline{2-17}
  & ACC & ASR
  & ACC$\uparrow$ & ASR$\downarrow$
  & ACC$\uparrow$ & ASR$\downarrow$
  & ACC$\uparrow$ & ASR$\downarrow$
  & ACC$\uparrow$ & ASR$\downarrow$
  & ACC$\uparrow$ & ASR$\downarrow$
  & ACC$\uparrow$ & ASR$\downarrow$
  & ACC$\uparrow$ & ASR$\downarrow$ \\
  \hline
    Badnet & 62.99 & 99.89 & 56.97 & 99.26 & 57.43 & 57.42 &  61.63 & 0.85 & 62.86 & 1.84 & 63.05 & 3.93 & 58.23 & 8.25 & 63.23 & 1.65\\
    Blend &63.30 & 99.70 & 59.98 & 1.02 & 59.84 & 62.17 & 61.94 & 11.55 & 62.20 & 3.96 & 62.49 & 0.61 & 60.21 & 4.23 & 62.37 & 1.19\\ 
    SIG & 64.10 & 99.74 & 59.13 & 42.13 & 58.62 & 40.68 & 62.02 & 9.60 & 61.34 & 4.13 & 63.85 & 4.56 & 60.01 & 7.41 & 62.58 & 4.12\\ 
    WaNet & 60.76 & 99.92 & 57.96 & 77.45 & 53.86 & 43.70 & 56.42 & 27.79 & 58.34 & 34.21 & 54.82 & 16.98 & 57.98 & 6.41 & 60.52 & 5.32 \\ 
\bottomrule
  \end{tabular}
 }
\label{tinyimagenet}
\vspace{-1.0em}
\end{table*}

\begin{figure*}[t]
\centering
\subfloat[Blend Attack (all-to-one).]{
\label{cm:1}
\includegraphics[width=0.24\linewidth]{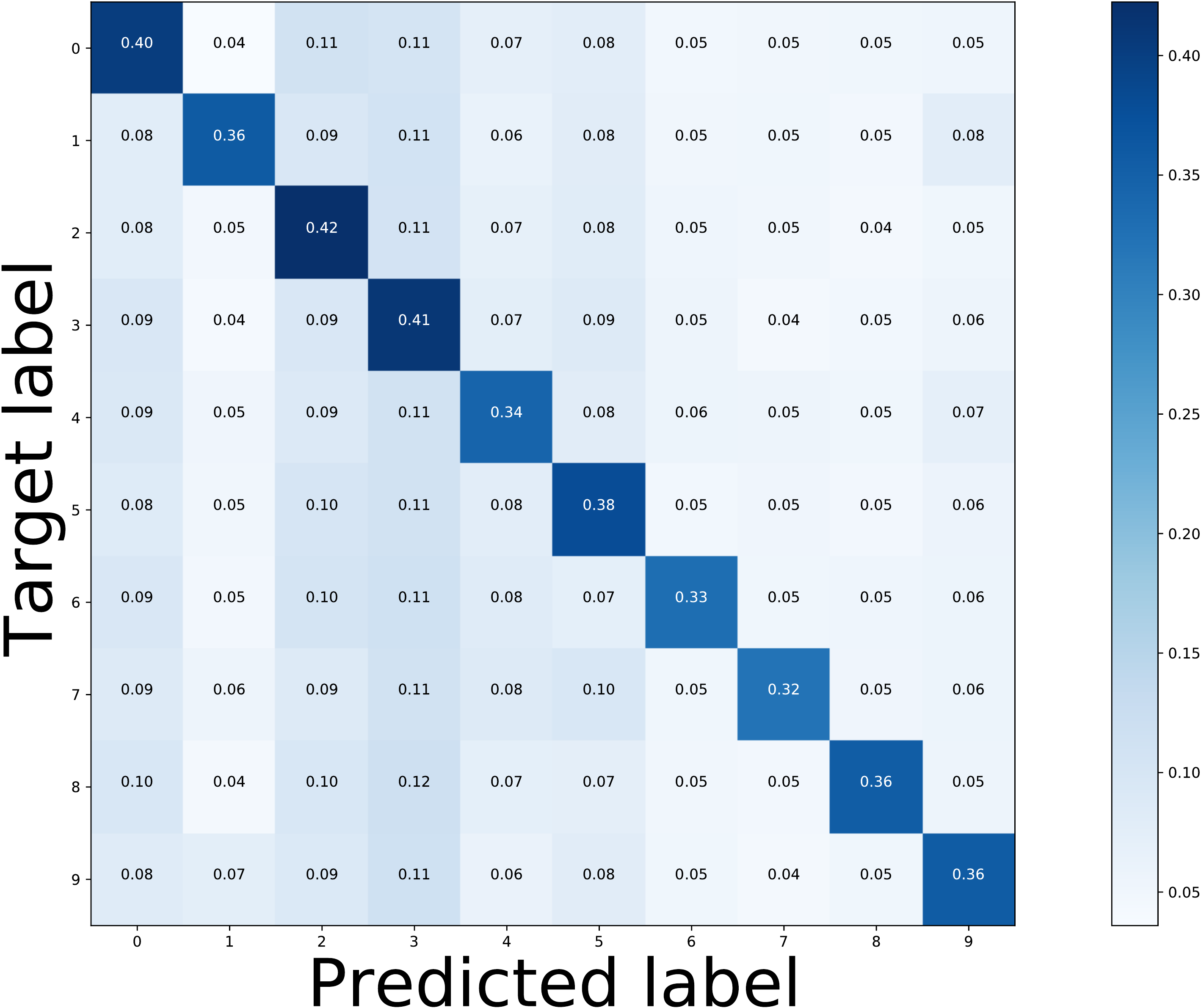}
}
\subfloat[SIG Attack (all-to-one).]{
\label{cm_more_1}
\includegraphics[width=0.24\linewidth]{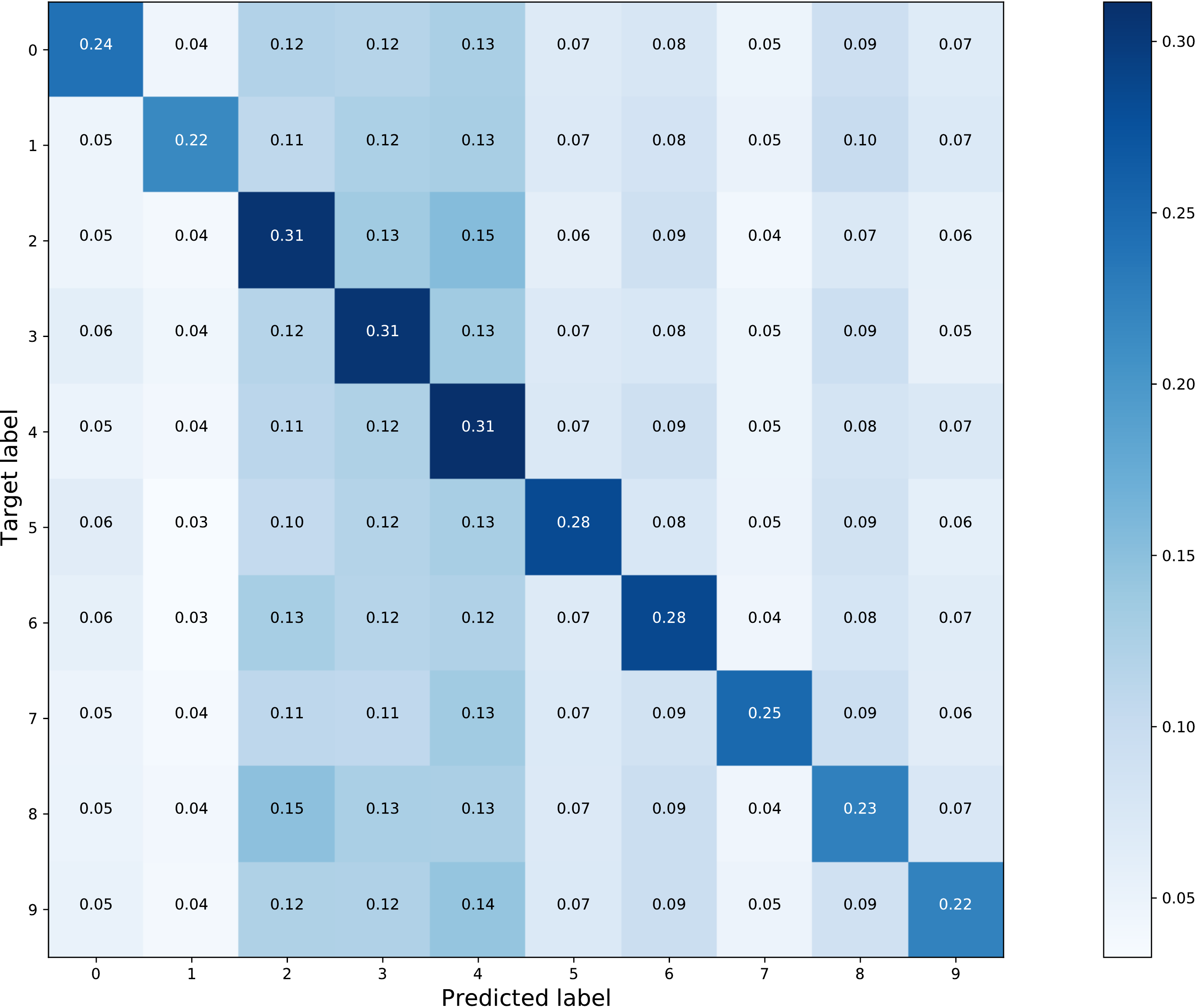}
}
\subfloat[WaNet Attack (all-to-all).]{
\label{cm_intro:3}
\includegraphics[width=0.24\linewidth]{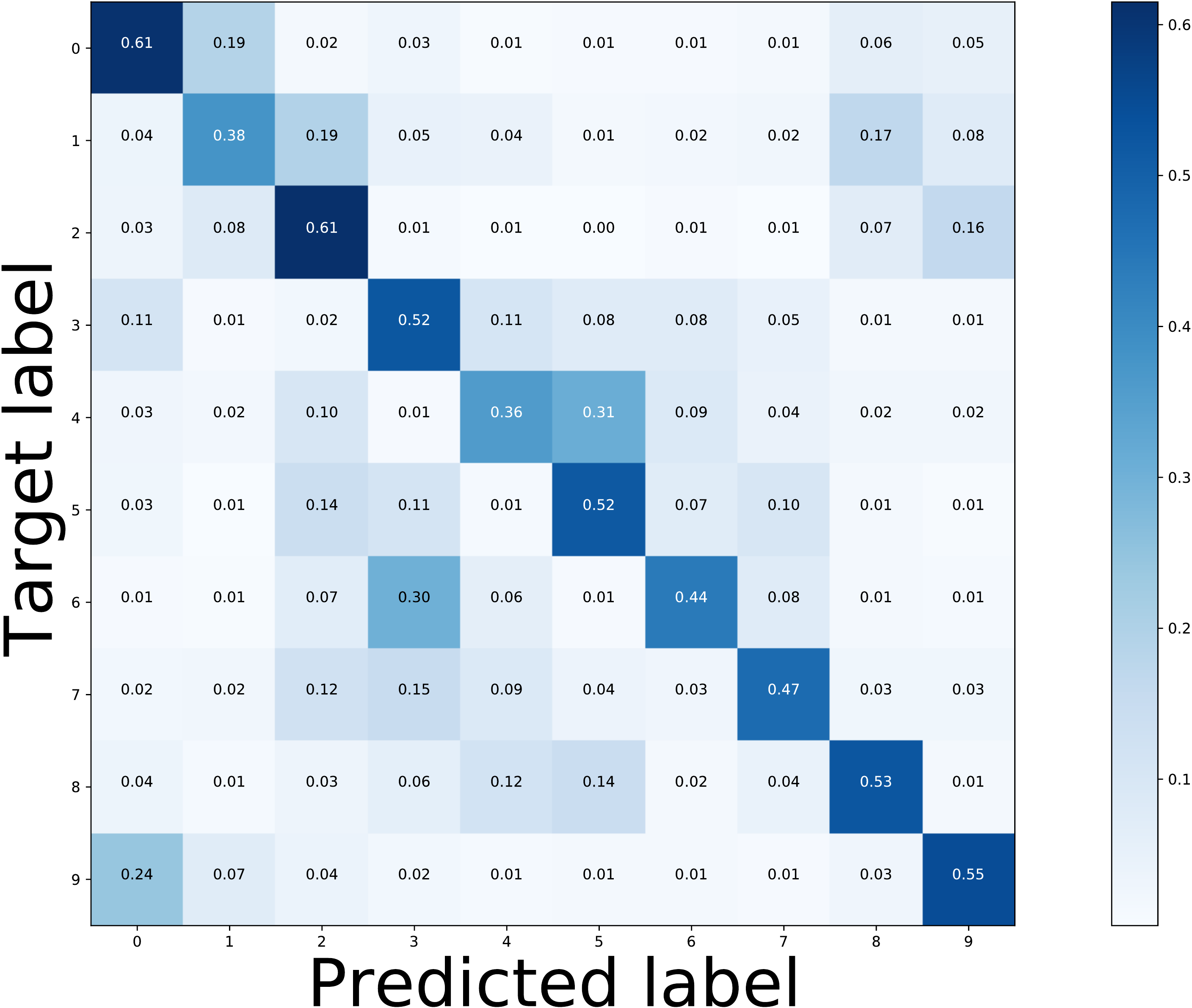}
}
\subfloat[GTSRB-sub. (all-to-one).]{
\label{cm:2}
\includegraphics[width=0.27\linewidth]{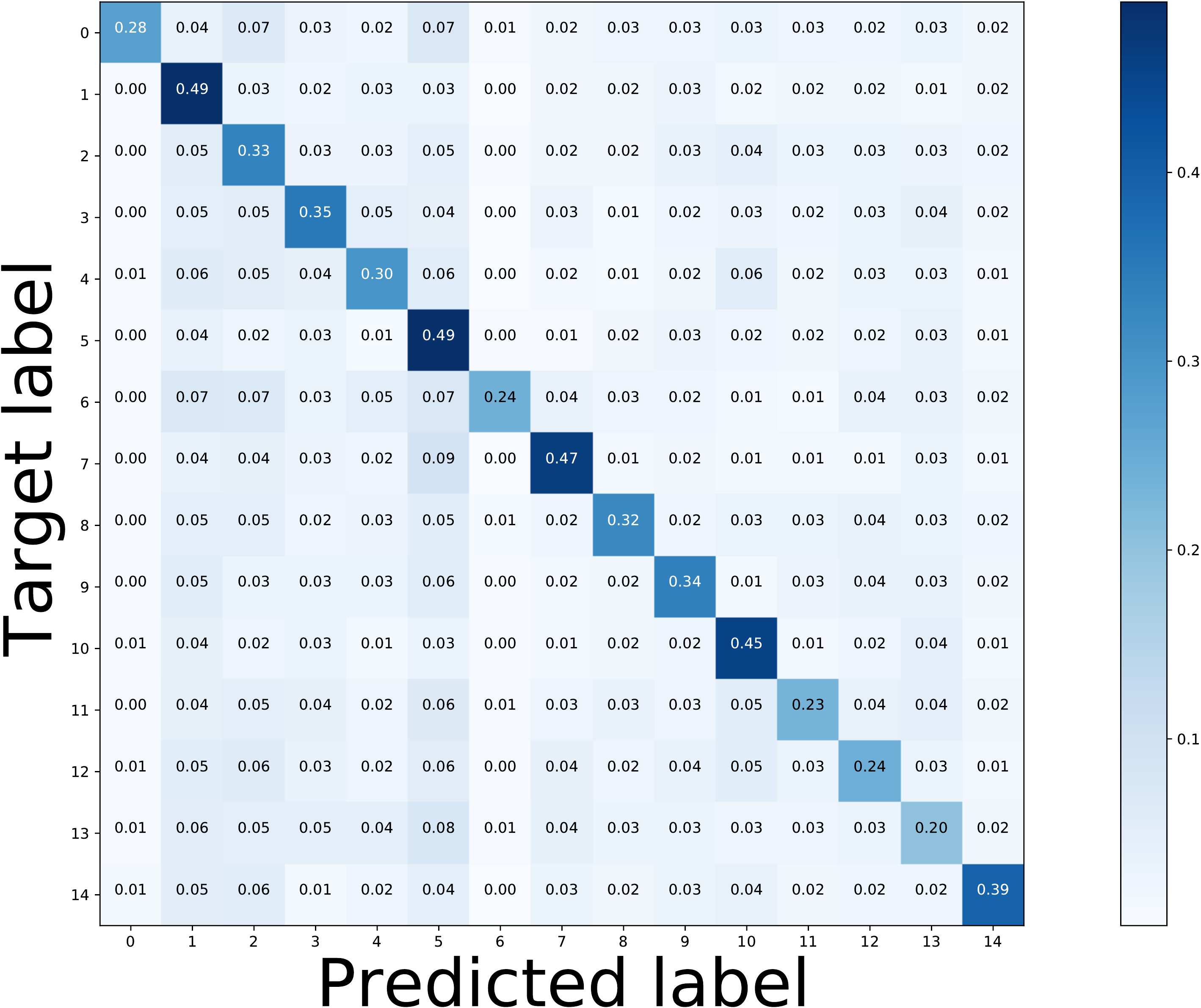}
}
\caption{Predicated labels v.s. Target-labels for adversarial examples from CIFAR-10 and GTSRB. No matter what attack methods are (\emph{e.g.,} Blend, SIG, WaNet), what attack settings are (\emph{e.g.,} all-to-one, all-to-all), what datasets are (\emph{e.g.,} CIFAR-10, GTSRB), the dominate predicted labels always align to the target-labels, as shown by the diagonal of the matrix.}
\vspace{-1.0em}
\label{cm}
\end{figure*}

\begin{table*}[t]
\scriptsize
 \centering
 \caption{Comparison with SoTA defense methods (at \textbf{all-to-all setting}) on CIFAR-10 dataset.}
 \vspace{-0.5em}
 \resizebox{1.0\textwidth}{!}{
 \setlength{\tabcolsep}{0.9em}%
  \begin{tabular}{c|c c|c c|c c|c c|c c|c c|c c|c c}
  \hline
  \multirow{0}* & 
  \multicolumn{2}{c|}{\textbf{Before}} & \multicolumn{2}{c|}{\textbf{Fine-tuning}} & 
  \multicolumn{2}{c|}{\textbf{Fine-pruning}} & \multicolumn{2}{c|}{\textbf{NAD}} & 
  \multicolumn{2}{c|}{\textbf{Neural Cleanse}} & \multicolumn{2}{c|}{\textbf{ANP}} & 
   \multicolumn{2}{c|}{\textbf{PBE (w/o clean)}} & 
  \multicolumn{2}{c}{\textbf{PBE (w/ clean)}} \\
  \cline{2-17}
  & ACC & ASR
  & ACC$\uparrow$ & ASR$\downarrow$
  & ACC$\uparrow$ & ASR$\downarrow$
  & ACC$\uparrow$ & ASR$\downarrow$
  & ACC$\uparrow$ & ASR$\downarrow$
  & ACC$\uparrow$ & ASR$\downarrow$
  & ACC$\uparrow$ & ASR$\downarrow$
  & ACC$\uparrow$ & ASR$\downarrow$ \\
  \hline
    Badnet & 94.63 & 94.41 & 85.54 & 3.68 & 88.20 & 66.13 & 89.63 & 1.01 & 94.63 & 94.41 & 92.01 & 0.69 & 93.62 & 0.68 & 93.84 & 0.62  \\
    Blend & 94.89 & 87.94 & 86.60 & 5.36 & 87.96 & 74.15 & 89.91 & 2.38 & 94.89 & 87.94 & 93.12 & 1.24 & 92.76 & 0.84 & 93.65 & 0.68 \\ 
    SIG & 94.66 & 84.34 & 87.98 & 2.83 & 88.99 & 69.52 & 91.53 & 1.36 & 94.66 & 84.34 & 93.60 & 0.87 & 93.71 & 1.07 & 93.52 & 1.01 \\
    DynamicAtt & 94.40 & 92.72 & 92.05 & 4.46 & 89.62 & 90.33 & 92.71 & 1.39  & 94.40 & 92.72 & 92.86 & 1.09 & 93.28 & 0.75 & 93.28 & 0.75 \\ 
    WaNet & 94.49 & 93.47 & 93.37 & 7.81 & 89.02 & 92.53 & 93.68 & 3.05 & 94.49 & 93.47 & 93.21 & 0.99 & 93.24 & 1.02 & 93.45 & 0.80 \\
\bottomrule
  \end{tabular}
 }
\label{all2all}
\vspace{-1.0em}
\end{table*}

\subsection{Comparison to SoTA Defense Methods}
In our experiments, all existing methods will use the clean extra dataset. In contrast, our approach could work either with or without such an extra dataset, both of them are reported in Table.\ref{tb1} and Table.\ref{tb2}.

From Table.\ref{tb1} and \ref{tb2}, if a clean extra dataset is available, our PBE defense can remarkably reduce ASR (\emph{e.g.,} down to $\textbf{1.09\%}$), meanwhile keep the ACC (\emph{e.g.,} at $\textbf{94.2\%}$). It indicates our approach outperforms other methods under the same condition (using a clean extra dataset). Besides, the Neural Cleanse (NC) cannot defend against the content-aware attacks (\emph{e.g.,} DynamicAtt, WaNet). It is due to that such trigger synthesis based methods need to recover a trigger, but content-aware attacks make triggers adaptive to image content, rather than using a fix trigger. 

In addition, if such a clean extra dataset is unavailable, all other defense methods \textbf{cannot} work, but our approach could still achieve excellent defensive performance (\emph{e.g.,} down to ASR=$1.16\%$) against invisible-trigger attacks (\emph{e.g.,} Blend, SIG, WaNet). Note that visible-trigger attacks (\emph{e.g.,} BadNet) can efficiently backdoor a model by using only several poisoning images. And our approach cannot perfectly filter out all poisoning images in the extra dataset, such that the defensive effect is a little weak (ASR=$11.3\%$).

Note that our approach could defend against SIG attack~\cite{barni2019sig}, which illustrates that our approach can effectively deal with clean-poisoning attack.

\vspace{0.4em}
\noindent\textbf{All-to-all Attack Setting:}
The previous comparisons are evaluated under the all-to-one attack setting, and we further evaluate our approach under the all-to-all attack setting. Following previous methods~\cite{nguyen2020wanet}, we set target-label as $y+1$. From Table.\ref{all2all}, it is obvious that our approach is also very effective in this attack setting. Note that Neural Cleanse has poor defensive performance for the all-to-all attack setting.

\vspace{-0.4em}
\subsubsection{Evaluation on Large-scale Dataset} To evaluate our approach on a large-scale dataset, we conduct experiments on the Tiny-ImageNet dataset~\cite{tinyimagenet}. The original Tiny-ImageNet contains $100,000$ images of 200 classes downsized to 64×64 colored images. In order to reload the pre-trained weights associated to the standard ImageNet dataset~\cite{deng2009imagenet}, we resize those images back to 224×224, and adopt Resnet-18 as the baseline classification model. From Table.\ref{tinyimagenet}, We can see that our approach is still effective on a large-scale dataset.

\begin{table}[t]\small
\vspace{-0.5em}
\caption{Quantitative comparisons for feature similarity.}
\vspace{-0.5em}
\centering
\resizebox{0.4\textwidth}{!}{
  \begin{tabular}{c | c | c | c |c | c}
  \toprule
    & Badnet &  SIG & Blend & DynamicAtt & WaNet \\
    \midrule
    $D_{\text{benign}}$ & 102.58 & 135.91 & 124.22 & 40.42 & 48.13 \\
    $D_{\text{infected}}$ & 85.11 &  78.18  & 75.09 & 28.66 & 15.85 \\
\bottomrule
  \end{tabular}
  }
\label{feature_similarity} 
\end{table}

\subsection{More Results for Our Observation}
\subsubsection{Predicted Labels v.s. Target-labels} Fig.\ref{cifar_hist} has illustrated one example of our observation that the adversarial examples are highly likely to be classified as target-label, under the condition of WaNet attack with all-to-one setting for CIFAR-10 dataset. In this section, we illustrate that such observations are present regardless of what attack methods are (\emph{e.g.,} Blend, SIG, WaNet), what attack settings are (\emph{e.g.,} All-to-one and All-to-all), and what datasets are (\emph{e.g.,} CIFAR-10, GTSRB and Tiny-ImageNet).

Finally, we observe similar trends, \emph{i.e.}, the dominant predicted labels always align to the target-labels, as shown by the diagonal of the matrix in Fig.\ref{cm}. Due to the limited space, for the GTSRB dataset we only randomly select $15$ from $43$ classes. More results for different configurations are provided in the supplemental material. 

\subsubsection{Comparisons of Feature Similarity} 
Fig.\ref{backdoor_feat} qualitatively illustrates that the features of an adversarial image $\widetilde{\bm{x}}'$ are very similar to that of the triggered image $\bm{x^t}$, rather than the clean image $\widetilde{\bm{x}}$. Here we conduct more quantitative comparisons. 

We randomly sample $10,000$ images from CIFAR-10, and calculate the $l_2$ distances between the features of $\widetilde{\bm{x}}'$ and $\bm{x^t}$, \emph{i.e.}, $D_{\text{infected}}=||f(\widetilde{\bm{x}}'),f(\bm{x^t})||_2$. Meanwhile, we also calculate the $l_2$ distances between the features of $\widetilde{\bm{x}}$ and $\bm{x^t}$, \emph{i.e.}, $||D_{\text{benign}}=f(\widetilde{\bm{x}}),f(\bm{x^t})||_2$. Regarding image features $f()$, we adopt the output of the last convolution layer (just before the fully-connected layer) as image features. From Table \ref{feature_similarity}, we can see that after planting a backdoor into a model, the feature distances $D_{\text{infected}}$ is smaller than $D_{\text{benign}}$ significantly.

\begin{figure}[htpb]
\centering
\includegraphics[width=0.95\linewidth]{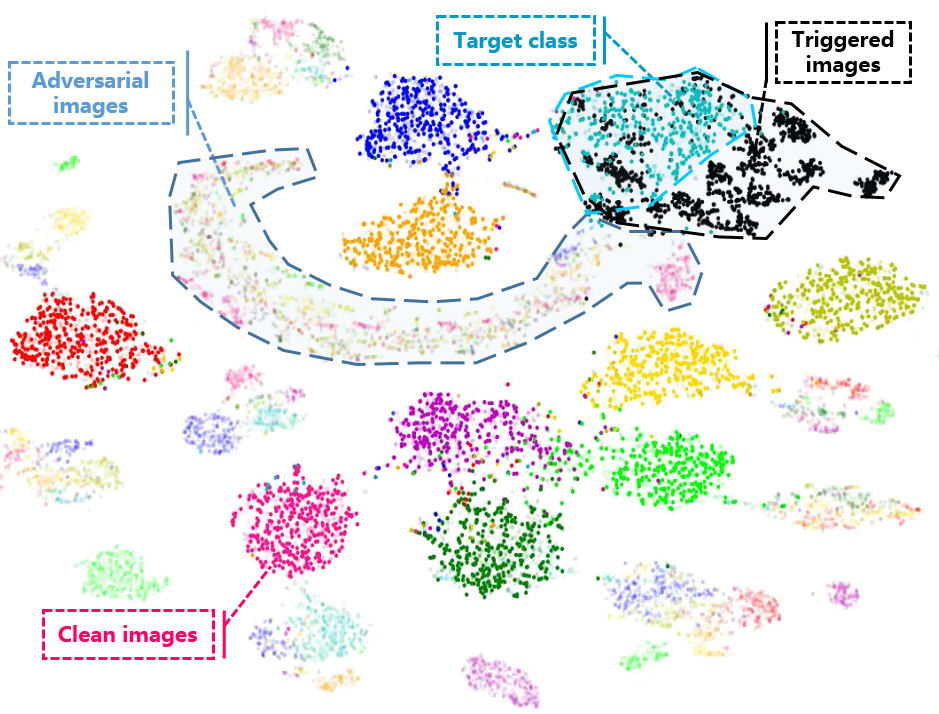}
\caption{The visualization of image features from CIFAR-10 dataset. The original images are in dark color (\emph{e.g.,} dark-red, dark-yellow, dark-green, \emph{etc}), while the corresponding adversarial examples are in light color (\emph{e.g.,} light-red, light-yellow, light-green, \emph{etc}). In this case, the target class is shown in cyan. The triggered images are shown in black, which is close to target class. Obviously, lots of adversarial examples lie on a `belt' which is close to the triggered images.}
\label{backdoor_tsne}
\end{figure}

Furthermore, we visualize those high-dimensional features in a 2D space with t-SNE. Fig.\ref{backdoor_tsne} shows the features of the original clean images in dark colors, their corresponding adversarial images in light colors, and the triggered images (in black). In this case, the backdoor target-label is shown in cyan. From Fig.\ref{backdoor_tsne}, it is obvious that most of adversarial images, triggered images, and target-label images lie on a same data manifold. It also justifies that the adversarial images are very similar to the triggered images.

\subsection{Progressive Learning}

\subsubsection{Identifying Clean Images}
In our approach, we formulate the clean image identification as an image ranking problem. Thus, we can evaluate its performance by using Precision-Recall curve and the Average Precision (AP) score. 
Specifically, for all training images $\bm{x}_i \in D_{train}$, suppose we know which one is poisoned, which is regarded as ground-truth. Then, we rank them in descending order according to their prediction changes $S_{\theta', \theta^t}(\bm{x})$ as Eq.\eqref{eq:idclean}. In our approach, the more top an image is ranked, the most possible it is regarded as a clean image. Therefore, we can evaluate our approach by comparing our ranking results to the ground-truth ranks. 

More importantly, our approach has an iterative procedure, which will \textbf{gradually improve the quality of purified model and extra dataset}. In practice, at the first iterations the infected model is not well purified, and hence we just select top ranked $10\%$ or $20\%$ training data as extra dataset. With improvement of the model purification, the quality of our clean image ranking and identification is also improved, and we will fetch more data into extra dataset (\emph{i.e.}, top ranked $70\%$ images).

\begin{figure}
\vspace{-1.0em}
\centering
    \includegraphics[width=7cm]{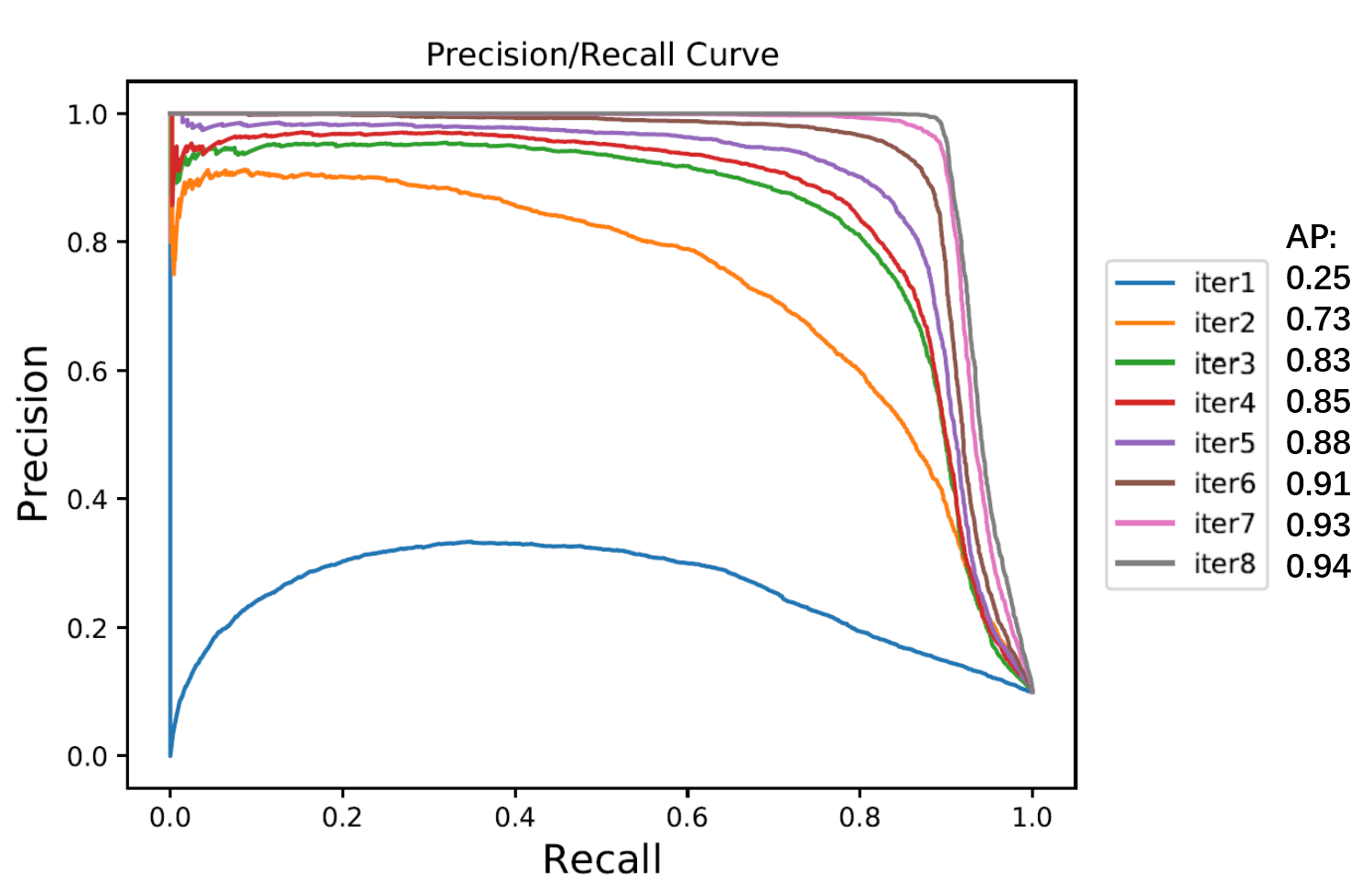}
\vspace{-0.5em}
\caption{The progress of clean-image identification with respect to the increase of iterations.}
\label{fig:pr}
\vspace{-0.5em}    
\end{figure}

Fig.\ref{fig:pr} is an example for our approach to defend against blend attack on CIFAR-10 dataset. Obviously, with the increase of iterations, the precision-recall curve becomes more and more better. It indicates that the quality of clean image identification is progressively improved. Beside, the corresponding Average precision (AP) raises from $0.25$ to $0.93$ gradually. 

\subsubsection{Progress of Purified Model}
\begin{figure}
\vspace{-0.5em}
\centering
\includegraphics[width=5.5cm]{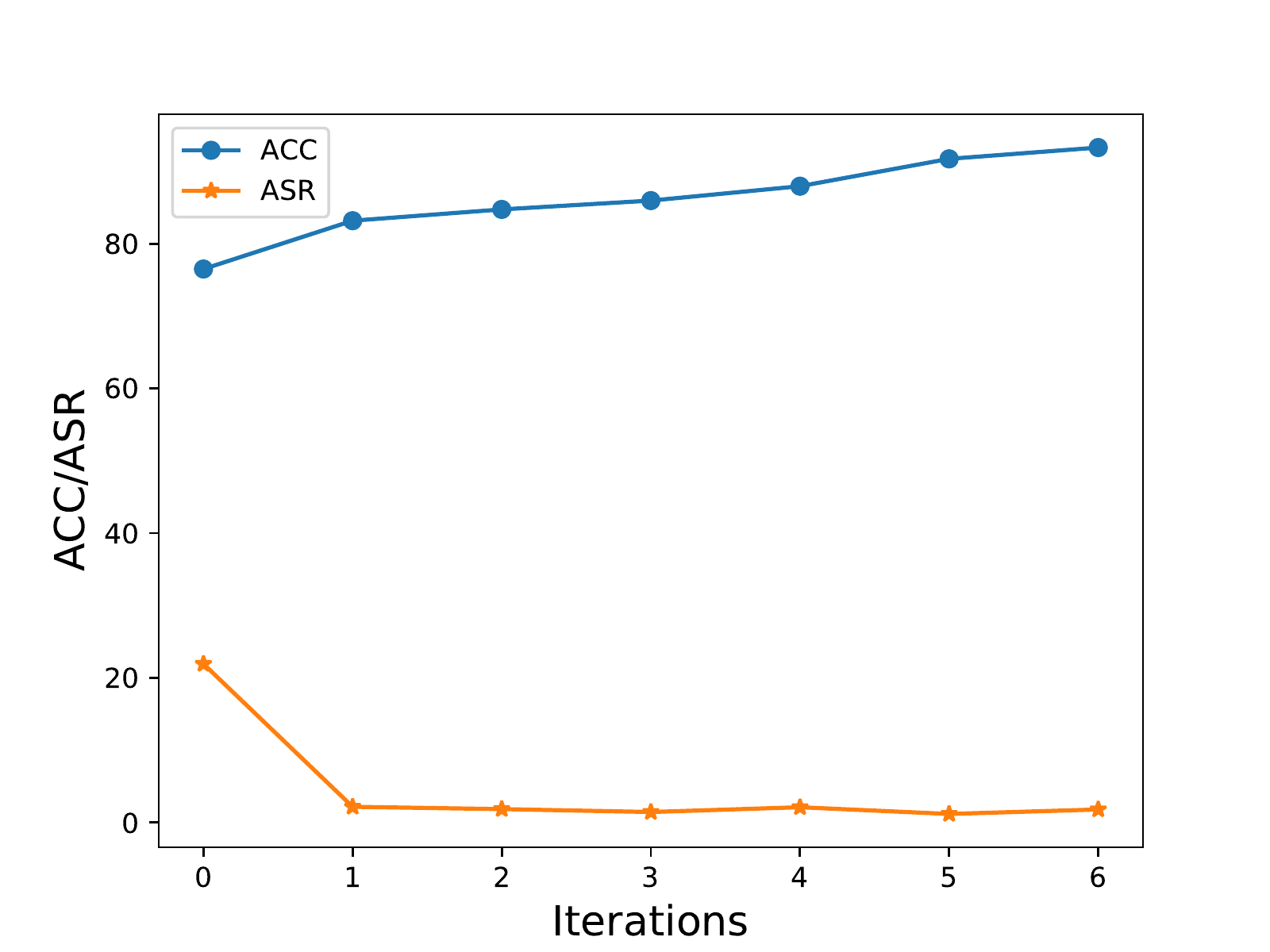}
\vspace{-0.5em} 
\caption{The progress of purified model with respect to the increase of iterations.}
\label{fig:it}
\vspace{-0.5em}    
\end{figure}

Meanwhile, the quality of purified model is also improved with the increase of iterations. From Fig.\ref{fig:it}, we can see that with the increase of iterations the ASR drops and ACC raises gradually, which indicates that the purified model is improved better and better on both benign and poisoning images. Particularly, the backdoor can be quickly erased (reducing ASR quickly) at beginning iterations, while the following iterations mainly help to improve the performance on benign images (raising ACC gradually).

\section{Conclusion}
In this work, we propose a new progressive backdoor erasing approach by leveraging adversarial attack techniques. It stems from our intriguing observations that for an infected model, its adversarial examples have similar behaviors as the triggered images. And an theoretical analysis is given to explain such observations. Importantly, such an underlying connections between adversarial and backdoor attacks will encourage our community to study them jointly.
In addition, our defense method could effectively defend against modern strong backdoor attacks (\emph{e.g.}, DynamicAtt, WaNet), even when a clean extra dataset is unavailable.

{\small
\bibliographystyle{ieee_fullname}
\bibliography{egbib}
}

\end{document}